\documentclass[12pt,twocolumn]{article}
\usepackage{graphicx}

\begin{document}

\title{Quality Estimation of Machine Translated Texts based on Direct Evidence from Training Data}
\author{Vibhuti Kumari \and Kavi Narayana Murthy\\School of Computer and Information Sciences,University of Hyderabad\\vibhuroy9711@gmail.com,knmuh@yahoo.com}
\date{}

\maketitle

\bibliographystyle{plain}

\begin{abstract}

Current Machine  Translation systems  achieve very  good results  on a
growing variety of  language pairs and data sets.  However,  it is now
well known that they produce fluent translation outputs that often can
contain important  meaning errors. Quality Estimation  task deals with
the  estimation  of quality  of  translations  produced by  a  Machine
Translation  system without  depending on  Reference Translations.   A
number of approaches have been suggested over the years. In this paper
we show  that the parallel corpus  used as training data  for training
the  MT  system holds  direct  clues  for  estimating the  quality  of
translations produced by the MT system. Our experiments show that this
simple  and direct  method  holds promise  for  quality estimation  of
translations produced  by any  purely data driven  machine translation
system.
  
\end{abstract}

\section{Introduction}

The performance of Machine Translation (MT) systems is measured either
using   Manual  evaluation   using  metrics   such  as   Adequacy  and
Comprehensibility, or  using automatic  methods using metrics  such as
BLEU and TER, by comparing with Reference Translations \cite{kaushal}.
Quality  Estimation (QE),  on  the other  hand,  deals with  automatic
estimation of quality of translations produced by an MT system without
using reference Translations.\\

QE  of MT  outputs has  several  benefits.  Good  translations can  be
selected, post-edited as required and added to the training data. Poor
quality cases  can be removed from  training data to reduce  noise. QE
helps in more accurate estimation  of post-editing time and effort and
in taking associated decisions in commercial translation.\\

A  large  number   of  techniques  have  been   proposed  for  quality
estimation.  The annual workshop on Machine Translation (WMT) has been
including  a shared  task on  quality estimation  for many  years now.
Chrysoula  Zerva et  al  \cite{zerva-etal-2022-findings} describe  the
findings  of the  11th edition  of this  shared task  held as  part of
WMT-2022. Participants  from 11  different teams  submitted altogether
991  systems  to  different  task   variants  and  language  pairs  in
WMT-2022.\\

Machine  translation  generally works  sentence  by  sentence and  the
primary goal of the Quality Estimation (QE) task is also to measure of
the quality of translations at  sentence level.  Several sub-tasks and
related tasks are  also taken up in the WMT  workshops.  Word level QE
deals with  marking of words  as OK or  BAD.  In fact,  sentence level
scores are often computed or  estimated using these word level scores.
Scoring entire documents  is another task.  Identifying  SL words that
cause  quality issues  is also  looked  at.  Explainable  QE task  and
Critical  Error   detection  task   were  included  in   the  WMT-2022
conference. Both Direct Assessment  on post-edit data (called MLQE-PE)
and  Multidimensional  Quality Metrics  (MQM)  were  included. In  the
current evaluation practices, QE systems  are assessed mainly in terms
of their correlation with human judgements.\\

Anna  Zaretskaya  et   al  \cite{zaretskaya-etal-2020-estimation}  ask
whether   the   current  QE   systems   are   useful  for   MT   model
selection. Serge Gladkoff et al \cite{Gladkoff} focus on the amount of
data that is required to reliably  estimate the quality of MT outputs.
They use  Bernoulli Statistical Distribution Modeling  and Monte Carlo
Sampling  Analysis   towards  this  end.  Shachar   Don-Yehiya  et  al
\cite{Shachar}  focus on  quality  estimation  of machine  translation
outputs in advance. They present a new task named PreQuEL, the task of
predicting the quality of the output of MT systems based on the source
sentence only. Some have focussed  on data set generation, others have
developed tools for QE.  While the research in MT QE  is rich in terms
of ideas,  techniques, tools, and  resources, it appears that  none of
them are looking  at the parallel corpus that is  used for building MT
systems  for clues  about quality  of translations.  In this  paper we
propose what we  call Direct Evidence approach, which  is based solely
on the training data that is used to build MT systems.

\section{Direct Evidence Approach}

Translation is  a meaning  preserving transformation  of texts  from a
Source Language (SL) to a Target Language (TL). This is generally done
sentence by sentence, or more  generally, segment by segment. In order
to preserve  the meaning of the  SL sentence, words and  phrases in SL
sentences need to be mapped to equivalent words and phrases in the TL.
Other aspects of  syntax and semantics such as  agreement, word order,
semantic compatibility will  also need to be  addressed. Modern purely
data driven  approaches such as Statistical  Machine Translation (SMT)
and Neural  Machine Translation (NMT) are  based on the view  that all
linguistic regularities  and idiosyncrasies are indirectly  present in
the parallel corpus and parallel  corpus alone is sufficient, no other
data or  linguistic resource  is needed.   A Machine  Translation (MT)
system can be obtained by training  on a training data consisting of a
parallel corpus alone.\\

We believe that the training data also has clues useful for estimating
the quality of translations produced  by the MT system. In particular,
here we focus on lexical transfer. We show that the Word Co-occurrence
Matrix (WCM) holds direct clues  for estimating the quality of lexical
transfer and hence quality of translation as a whole.\\

Statistical basis  for performing  lexical transfer comes  mainly from
word  co-occurrence  statistics.  Let   SL-TL  be  a  parallel  corpus
consisting of n Source Language segments $S_1,S_2,S_3,...,S_n$, paired
with  their  translational  equivalents $T_1,T_2,T_3,...,T_n$  in  the
Target Language. We say  SL word i co-occurs with TL word  j if the TL
word  j  occurs anywhere  in  the  translational  equivalent of  a  SL
sentence in which  the word i occurs.  Let $V_s$  be the Vocabulary of
the Source Language (total number  of distinct word forms occurring in
any of  the SL  segments) and  $V_t$ be the  Vocabulary of  the Target
Language. Then Word Co-Occurrence Matrix WCM is a $V_s$ x $V_t$ matrix
of non-negative integers where  $WCM_{i,j}$ indicates the total number
of times  the Source Language word  i had co-occurred with  the Target
Language word j in the entire  training data set. Clearly, WCM will be
a very large and very sparse matrix.\\

A large $WCM_{i,j}$  value indicates a strong  correlation between the
SL word  i and  TL word  j in  the training  corpus. If  an SL  word i
co-occurs with a TL word j large  number of times, if i does not occur
with too  many other TL words  with high frequency, if  the WCM counts
for other possible  mappings in TL are significantly  lower, all these
indicate that the lexical transfer of i to j during translation can be
done  with  high  confidence.  When   the  evidence  in  the  form  of
co-occurrence counts  coming from  the training data  is weak,  the MT
system may still go  ahead and substitute the word j  for word i based
on  the combined  evidence coming  from other  parts of  the sentence,
language model, etc.  This may be an optimal decision  taken by the MT
system with regard to some  specified loss function. Optimal choice in
some probabilistic sense may not be the correct choice, it may just be
the best of several possible choices, none of which may be correct. MT
systems generally go ahead and  produce translations, whether they are
sure or not-so-sure or not-at-all-sure.\\

Here we hypothesize  that the fraction of words in  a SL sentence that
have strong  co-occurrence relations with any  of the words in  the TL
sentence produced by  the MT system, is a direct  indicator of quality
of translation.

\section{Experiments and Results}

In  our first  experiment we  use an  English-Kannada parallel  corpus
consisting  of 4,004,894  segments  (that is  approximately 4  Million
segments) \cite{ramesh2021samanantar}  There are  about 36M  tokens in
English and 27M tokens in Kannada.  The Vocabulary size for English is
281,881. Only 42,222 (less than 15\%)  occur at least 20 times. 78.5\%
of words  occur less than  10 times, 69\% of  words occur less  than 5
times,  44.47\%  of  words  occur   only  once.   This  highly  skewed
distribution of words  in all human languages is  very well understood
and  expressed  through  laws  such  as  Zipf's  law  \cite{zipf}  and
Mandelbrot's law \cite{mandelbrot}. The Vocabulary of the Kannada part
is 1,253,589.   This number  is larger  due to  the much  more complex
morphology we see in Dravidian  languages such as Kannada. Only 82,227
(6.5\%) occur at  least 20 times.  89.2\% of words  occur less than 10
times, 81.8\% of words occur less  than 5 times, 57.8\% of words occur
only  once. The  general  picture  will be  similar  for  any pair  of
languages in the world.\\

If a SL word i occurs only once and the translation of the sentence in
which it occurs has n words, then i  can be mapped to any one of these
n TL words  with equal probability.  While an MT  system may use other
clues such as mappings of other  words in the SL sentence and language
model probabilities,  it will still be  decision that is not  based on
very  strong evidence.  Low  frequency words  show poor  co-occurrence
relations   and   hence   less  statistical   evidence   for   lexical
transfer. Low frequency words are large  in number in any language and
this is a big  issue for any purely data driven  model. Larger data is
better but whatever  may be the size of the  data, the problem remains
pertinent.\\

Very  high frequency  words  can also  pose  challenges. They  usually
include determiners, prepositions and other function words. Words such
as 'the', 'of',  'by' occur with very high frequency  in English, none
of them  map to  any word in  Kannada. WCM will  show large  number of
possible mappings, all (or almost all)  of them will be wrong. This is
again  a hopeless  situation.   Phrase based  approaches and  sub-word
models attempt  to address these  problems and are successful  to some
extent.\\

Keeping these ideas  in mind, we build WCM for  words that co-occur at
least 20 times in the training  set, we exclude words which occur more
than 10,000 times in the  corpus.  Under these assumptions, WCM matrix
can be built very fast (it took less than 4 minutes on a 40 core Intel
Xeon Silver 4114  CPU at 800 MHz server) and  the size of uncompressed
the WCM  file is only  44 MB. There are  1,474,792 entries in  the WCM
matrix, there are only 38,502 English words in this matrix.\\

We divide  the corpus  into training, development  and test  sets with
4,004,894, 5000 and 5037 segments respectively and train an SMT system
using  MOSES \cite{koehn-etal-2007-moses}.  WCM  is  computed for  the
training set.\\

Then for each  segment in the test  set, we check the  number of words
(excluding  very  high frequency  words)  for  which there  is  strong
evidence in the training  data. This we do by checking  if the SL word
co-occurs at least 20 times with any of the TL words in the translated
text. We take the percentage of  words with strong evidence as a score
for ranking  the translations.  We  call these scores  Direct Evidence
(DE) Scores. DE Scores range from 0 to 100.\\

We run the trained  SMT system on test data. We  compute the DE Scores
as described above for each segment.  We pick out SL-TL pairs from the
test data  as also  from the  generated MT  outputs based  on selected
ranges  of  DE  Scores.  Taking  the  TL  part  in  the test  data  as
Reference, we compute BLEU scores: See Table 1.\\

\begin{table}
\begin{tabular}{|l|r|r|}
\hline 
DE Score & No. of Segments & BLEU Score\\
\hline
$<$ 20   &    847 &  6.33\\
$<$ 30   &   2082 &  6.78\\
$<$ 40   &   3036 &  7.06\\
$<$ 50   &   3588 &  7.46\\
$\ge$ 50 & 1449   &  9.16\\
$\ge$ 60 &  669   & 10.49\\
$\ge$ 70 &  327   & 11.34\\
$\ge$ 80 &  237   & 10.80\\
\hline
\end{tabular}
\noindent\caption{DE Scores vs. BLEU Scores for English-Kannada}
\end{table}

We can  clearly see a  positive correlation  between the DE  Scores we
obtained  and  the BLEU  scores,  up  to  a  threshold of  70.  Manual
observations  also   clearly  showed  the  gradation   in  quality  of
translations  correlating with  the  DE Scores  we compute.  Sentences
which got  high DE  Scores were generally  of much  better translation
quality compared to sentences which got a poor DE Score.\\

Next we  compute sentence level  BLEU scores and look  for correlation
between these  BLEU scores and  the DE  Scores. Over 5037  segments of
test data, we get a  Pearson Correlation Coefficient of 0.209405.  The
p-value is $<$ 0.00001 Hence the  result is significant at the typical
p $<$ 0.05.\\

Training corpora used for building  MT systems are often not available
for us to  experiment with.  Here we  take up one case  where we could
locate  the  training  data  as  also the  MT  outputs  and  Reference
Translations.  This  relates to English-Hindi SMT  system developed by
Piyush Dungarwal et al  from IIT Bombay \cite{dungarwal-etal-2014-iit}
in the  Ninth Workshop  on SMT, WMT-2014.   Training data  consists of
273,885  segment  pairs, including  3,378,341  tokens  in English  and
3,659,840  tokens in  Hindi. There  are 129,909  unique word  forms in
English, of which only 19,100 occur 10 times or more.  Total number of
unique word forms  in Hindi is 137,089, of which  only 18,587 occur 10
times or  more. In English,  30 words  occur with frequency  more than
10,000 and are  taken as frequent words in our  experiments. In Hindi,
there are  33 very high  frequency words.  These high  frequency words
are excluded from WCM computations.  This makes the WCM matrix smaller
and saves time too. Also, very  high frequency words co-occur with too
many words in TL and the  evidence for proper lexical transfer becomes
blurred. The  WCM matrix  could be computed  in a minute  or so  on an
ordinary  Desktop  computer.   The  WCM  has  642,341  entries.   This
includes 242,477 pairs that co-occur 20 times or more.\\

There are  2507 segments  in each  of the test  set source,  MT system
output, and  Reference Translations.  We compute  the DE  Scores based
purely on the  WCM matrix which is based only  on the training corpus.
We  extract subsets  of  the MT  outputs  and corresponding  reference
translations based  on the  DE Score  ranges. The  BLEU scores  are as
shown in Table 2.\\

\begin{table}
\begin{tabular}{|l|r|r|}
\hline 
DE Score & No. of Segments & BLEU Score\\
\hline
$<$   50 &  133 &  6.00\\
$\ge$ 50 & 2374 & 10.36\\
$\ge$ 60 & 2154 & 10.61\\
$\ge$ 70 & 1733 & 10.93\\
$\ge$ 80 & 1120 & 11.69\\
$\ge$ 90 &  463 & 12.47\\
\hline
\end{tabular}
\noindent\caption{DE Scores and BLEU Scores for English-Hindi}
\end{table}

Here again  we see a clear  gradation in BLEU scores  correlating with
the DE Scores. Higher the DE Score, higher the BLEU.\\

The results  of these preliminary  experiments support our  claim that
the clues  needed for MT QE  are present in the  training data itself,
nothing else may  be necessary.  We do  not even need an  MT system to
predict the quality of translations it will produce, just the training
data is sufficient.\\

We then  calculated the DE-Scores  for the 4 Million  segment Training
Data used for building our  English-Kannada SMT system. Figure 1 shows
the histogram  plot of DE-Scores obtained.  It can be observed  that a
significant part of the training data got DE-Scores less than 50, many
cases even less than  10. This can be useful in locating and  reducing
noise in the training data.\\

\begin{figure}[h]
  \includegraphics[width=0.99\columnwidth]{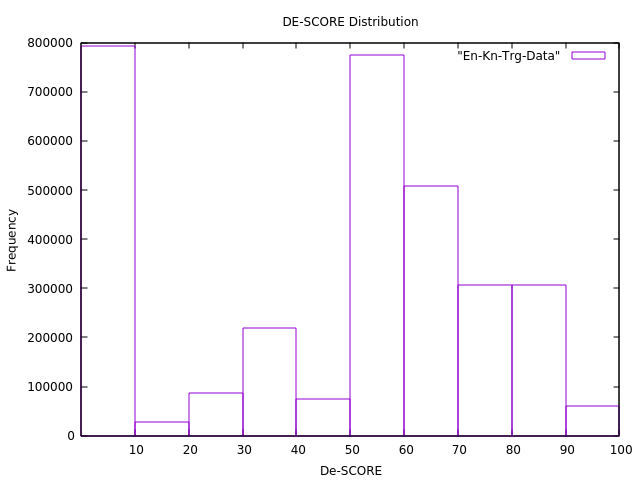}
  \caption{DE-Scores for English-Kannada Training Data Set}
\end{figure}

\section{Conclusions}

In  this  paper we  hypothesize  that  the  Parallel Corpus  used  for
Training an  MT system holds  clues about the quality  of translations
the MT system can produce.  We propose a simple and direct approach to
quality  estimation  based  solely  on  the  training  data.   A  word
co-occurrence matrix is constructed from  the training corpus and used
to estimate  the sentence  by sentence  quality of  translations. Each
sentence gets  a score  called DE  Score, which  is indicative  of the
quality of  translation.  Manual  observations show that  good quality
translations generally tend  to get higher DE Scores  and poor quality
translations  tend to  get  lower scores.   Our experiments  reconfirm
this.  This  simple  and  direct   evidence  approach  to  MT  Quality
Estimation appears  to holds promise.  We can estimate the  quality of
translations even  without / before running  the MT system. We  do not
need any other data or resource, we only need the training corpus.\\

We have used raw frequency  counts and manually selected thresholds to
decide which SL words have  sufficient evidence in the training corpus
for reliable lexical  transfer. Instead of counting  the percentage of
words in the  SL sentence which have enough evidence  (as indicated by
the co-occurrence counts),  we could use the  actual counts themselves
to get a more fine grained picture.  We could check how many and which
words in TL sentence co-occurred how many times with each of the words
in the SL sentence. We could look  at the frequency counts for all the
TL words that co-occur with a  given SL word, which particular TL word
has contributed in the given sentence  pair, which other TL words have
higher or lower frequency counts, how far is the next more frequent or
next less frequent TL word in  the WCM matrix and so on. Co-occurrence
in shorter sentences is more  significant than co-occurrence in longer
sentences   and  this   could  also   be  factored   into  the   score
computation.\\

DE-Scores provide us  a spectrum of quality grades and  since they are
based on co-occurrence counts, Out  of Vocabulary (OOV) words are only
cases that lie just outside the low end of this spectrum.\\

Missing words automatically get reflected  in poor DE Scores but extra
words in  TL can be detected  by performing a  TL to SL WCM  check. If
large scale manual  post-edit data such as HTER  scores are available,
then we  can estimate  the various  thresholds using  machine learning
techniques instead of using human judgement as we have done here.\\

In this work  we have only focused on only  one aspect, namely quality
of lexical transfer,  to judge the quality of  translations.  It needs
to  be explored  if  other  aspects such  as  agreement and  syntactic
completeness  and   correctness  of  dependency   relations,  semantic
coherence etc.  can  also be estimated from the  training corpus.  For
example, a  word co-occurrence  matrix built only  on the  TL segments
(where co-occurrence is defined as appearing in the same segment), may
be useful  in dealing  with agreement  and coherence  issues. Sub-word
models may be incorporated.\\

\bibliography{mtqe}

\end{document}